\documentclass[11pt]{article}

\usepackage[final]{acl}
\usepackage{array}
\usepackage{times}
\usepackage{booktabs}
\usepackage{latexsym}
\usepackage{amsmath}
\usepackage{multirow}
\usepackage[T1]{fontenc}

\usepackage[utf8]{inputenc}
\usepackage{microtype}

\usepackage{inconsolata}

\usepackage{graphicx}
\usepackage{xspace}
\newcommand{\methodname}{\textsc{OneModel}\xspace} 
\newcommand{\company}{Ant International\xspace}
\newcommand{\platform}{\texttt{Ant International Platform}\xspace} 
\usepackage{amssymb}

\title{How to Train a Real-World Silicon Concierge? Internalizing Complex Business Workflow to Only \methodname}

\author{Ant International}

\begin{document}
\maketitle
\begin{abstract}
Traditional industrial agents rely on modular pipelines, including Router, Retriever, Planner, Executor, Responder, Reviewer, etc., which inevitably fracture into a labyrinth of ad-hoc patches, leading to cascading errors and high latency. 
We propose \methodname, an applicable paradigm shift from external workflows to internalized knowledge representation. Unlike modular systems that slice fluid user intents into static steps, \methodname consolidates complex business logic and SOPs directly into the model’s parameters.
Through Continual Pre-training (CPT) and logic-compilation SFT, we transform fragmented business rules into the model’s intuitive reasoning within a unified attention space. Deployed in our global financial service system, \methodname effectively breaks the trade-off between latency, accuracy, and complexity. 
Online A/B testing demonstrates end-to-end latency reduction of more than 50\% (18.7s $\rightarrow$ 8s) while the Intelligent Resolution Rate (IRR) jumps from 64.3\% to 83.3\%. 
The results demonstrate our paradigm \methodname effectively replaces brittle engineering logic with internalized cognitive intuition, offering a scalable and future-proof blueprint for transitioning industrial agents from complex, error-prone workflows to unified model architectures.
\end{abstract}

\section{Introduction}

\begin{figure*}[htbp!]
    \centering
    \includegraphics[width=1\linewidth]{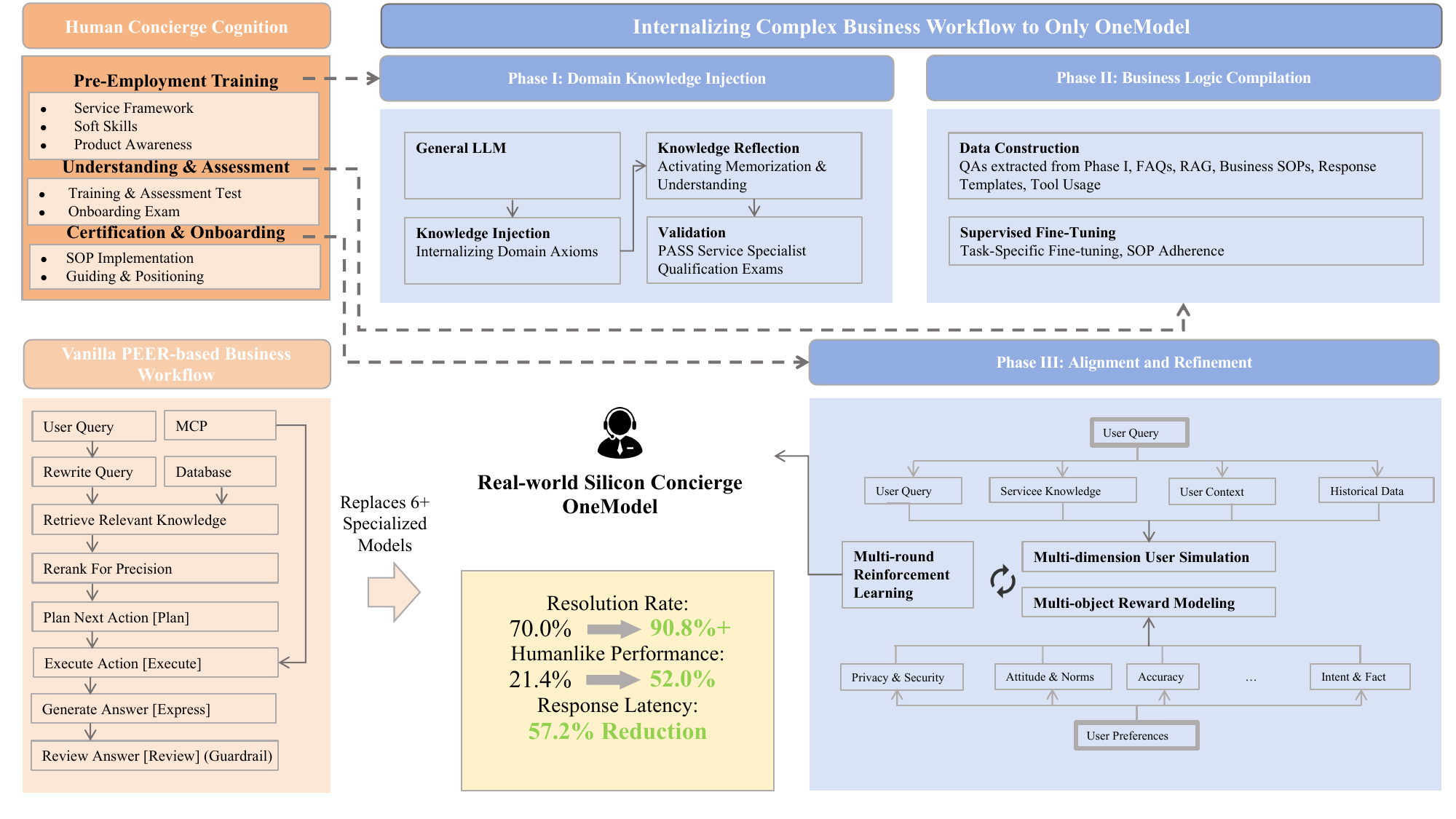}
    \caption{Architectural comparison between the traditional PEER-based modular workflow (left) and our proposed \methodname internalized knowledge pipeline (right). While PEER decomposes queries through six serialized stages, \methodname consolidates intelligence into a unified attention space and hence implicitly utilizes the models' evolving and emergent capabilities. Our online A/B testing has achieved 90\%+ precision with 50\%+ latency reduction (18.7s→8s) compared with traditional complex workflow.}
    \label{fig:overall_pipeline}
\end{figure*}

Traditional industry standard for complex business service has largely relied on modular and agentic workflows, such as \textit{Route--Retrieve--\textbf{P}lan--\textbf{E}xecute--\textbf{E}xpress--\textbf{R}eview} (PEER)~\citep{wang2024peerexpertizingdomainspecifictasks,hong2024metagptmetaprogrammingmultiagent,yao2023reactsynergizingreasoningacting,chen2025dingtalkdeepresearchunifiedmulti}. 
In practice, this architecture inevitably fractures into a labyrinth of ad-hoc patches. 
To mitigate specific bad cases, engineers are forced to implement intricate traffic-splitting logic, creating specialized, narrower sub-links to handle scenarios like ambiguous intent clarification, user counter-inquiries, or multi-modal inputs~\citep{huang2024largelanguagemodelsselfcorrect,press2023measuringnarrowingcompositionalitygap,tong2024llmslearnpreviousmistakes}. 
While such intricate pipelines theoretically offer fault isolation, they are increasingly becoming a bottleneck in the era of powerful foundational models.

This rigid fragmentation imposes a ceiling on intelligence.
By slicing fluid user intents into static, pre-defined sub-steps, most current architectures \textbf{fail to leverage the emergent capabilities inherent in modern or future foundational models}~\citep{dziri2023faithfatelimitstransformers}. 
Moreover, such complexity makes error attribution a major challenge: when the system fails, pinpointing the specific responsible module is difficult, making it extremely \textbf{difficult to incrementally refine the system}.
Crucially, the serial dependency of these modules \textbf{creates a risk of cascading error accumulation}: a minor inaccuracy in an upstream module (e.g., retrieval noise or intent misalignment) is amplified as it propagates downstream~\citep{yoran2024makingretrievalaugmentedlanguagemodels, tong2023eliminatingreasoninginferringplanning}.
In a modular system, the final generation model typically "blindly trusts" these retrieved contexts, leading to hallucinations that are confident yet factually incorrect~\citep{liu2023lostmiddlelanguagemodels}.
Finally, the serial execution of multiple pipelines might significantly \textbf{inflate Time-to-First-Token (TTFT)}, degrading the user experience precisely when speed is critical.

To solve these structural barriers and future-proof industrial deployments, we explore the \methodname paradigm, shifting from External Workflow to Internalized Knowledge~\citep{khattab2023dspycompilingdeclarativelanguage}. 
Deployed within our large-scale and high-stakes financial merchant service system, \methodname adopts a single-model architecture. 
By consolidating the full interaction loop—spanning intent reasoning, tool usage, and SOP adherence—into a unified attention space, this approach internalizes domain and enterprise knowledge as intrinsic parameters for direct generation.

However, injecting volatile data (e.g., exchange rates) directly into parameters is counterproductive, causing "knowledge obsolescence" and hallucinations due to the prohibitive cost of frequent retraining~\citep{lazaridou2021mindgapassessingtemporal,kandpal2023largelanguagemodelsstruggle}. To reconcile parameter rigidity with operational fluidity, we propose a Hierarchical Knowledge Management and Injection strategy. We strictly reserve Continual Pre-training (CPT) for static Fundamental Domain Principles to establish a robust semantic foundation. Semi-static Procedural Business Logic—including complex SOPs and response scripts—is compiled into intuitive reasoning via Supervised Fine-Tuning (SFT)~\citep{gururangan2020dontstoppretrainingadapt}. Meanwhile, Reinforcement Learning (RL) addresses error-prone nuances and robotic phrasing, reinforcing knowledge internalized by CPT and targeting long-tail failures beyond the reach of SFT. Finally, Volatile Transactional Context is decoupled via Dynamic Context Injection, ensuring real-time accuracy without the burden of retraining.


Extensive online A/B testing confirms \methodname effectively breaks the trade-off between latency, accuracy, and complexity. 
We also surprisingly find that \methodname can handle complex multi-turn inquiries with an organic, human-like touch that traditional rule-based or modular pipelines struggle to replicate.
Our successful deployment offers a compelling proof-of-concept for the whole industry: shifting focus from heavy workflow engineering to industrial model evolution is vital and critical for the future autonomous enterprise agents.

\section{Real-World Applications}

At \company, one of the leading global payment and financial companies, we deployed \methodname on the \textit{Global Merchant Service Agent}. 
This centralized system is mission-critical, automating thousands of intricate financial operations—ranging from cross-border payment inquiries to complex refund disputes and e-commerce settlements. 
Given the high stakes of pecuniary transactions, any deployment requires rigorous validation. 
Hence, we conducted extensive Online A/B Testing over multiple weeks, pitting our proposed \methodname against the existing modular baseline (a standard PEER pipeline). 
To holistically evaluate the transition, we thoroughly monitored three distinct dimensions: System Efficiency (Latency), Business Effectiveness (Resolution Rate), and Interaction Quality (Expert Annotation).

Notably, our 8B \methodname achieved a peak resolution rate of \textbf{90.75\%}, significantly outperforming commercial baselines including \textit{Claude-3.5-Haiku (86.72\%), Gemini-2.5-Pro (87.55\%), and GPT-5.2 (87.55\%)}.
In terms of business outcomes, the Overall Intelligent Resolution Rate (IRR)—defined as the percentage of user inquiries fully resolved by the AI agent without escalation to human staff—surged from \textbf{64.3\% to 83.3\%}.
Operationally, the unified architecture eliminated inter-module overhead, achieving a significant \textbf{57.2\% reduction} in latency—a critical improvement for impatient users facing financial uncertainties.

\section{Methods: The \methodname Paradigm}


To bridge the gap between fragmented modular workflows and the need for holistic reasoning, \methodname employs a hierarchical knowledge management and injection strategy. 
This approach categorizes business intelligence into three tiers based on volatility and complexity, internalizing them through a multi-stage training pipeline.

\subsection{Hierarchical Knowledge Management}
Unlike traditional RAG-heavy systems that treat all data as external evidence, our framework first differentiates domain knowledge into a hierarchy to optimize the trade-off between parameter rigidity and operational fluidity:

Fundamental Domain Axioms (Static): Core financial principles and multilingual semantics.

Procedural Business Logic (Semi-static): Complex SOPs and decision-making flows.

Volatile Transactional Context (Dynamic): Real-time data (e.g., exchange rates).

\subsection{Phase I: Knowledge Injection and Reflection}

To dismantle the reliance on brittle external retrieval—a common bottleneck in traditional modular workflows—we treat intrinsic domain expertise as a first-order requirement. We implement Continual Pre-training (CPT) to shift the model’s distribution from general text to the rigorous logic of our financial ecosystem. The primary objective is to equip a compact base model with "expert-level" closed-book proficiency, effectively mimicking the intensive onboarding and examination process of human specialists.

To achieve this, we adopt a two-stage internalization paradigm designed not just to memorize facts, but to transition domain-specific knowledge into actionable parametric memory: \textit{Knowledge Injection} and \textit{Knowledge Reflection}.

\paragraph{Stage 1: Knowledge Injection (KI) for Parametric Foundation.}
The initial stage focuses on constructing a dense, "full-stack" business knowledge background. Rather than indiscriminately feeding raw text, we aggregated a diverse, multi-tiered dataset comprising official business manuals, structured internal FAQs, and scrutinized online Q\&A "bad cases." By integrating both standard declarative rules and nuanced, real-world exceptions, we systematically eliminated critical blind spots in the model's understanding of business logic. Consequently, this phase transcends standard language modeling; it is a targeted knowledge injection that ensures the model's parametric memory is both deep and wide enough to support subsequent high-stakes reasoning tasks.

\paragraph{Stage 2: Knowledge Reflection (KR) for Memory Activation.}
However, establishing a memory reservoir is insufficient. Our findings indicated that raw parametric knowledge often remains "dormant"; without appropriate scenario-based stimulation, the model behaves as a passive information repository rather than an active problem solver. To transform dormant memory into active cognitive capabilities, we implemented a dual-stage activation pipeline using knowledge compression technique.

First, we utilized small-batch SFT as a verification probe to identify and map out high-impact business scenarios. These probes help us isolate the essential cognitive triggers for retrieval, memory, and reflection. In the subsequent synthesis stage, we leverage these verified scenarios as blueprints to reconstruct the raw pre-training corpus, ultimately synthesizing a large-scale, high-quality CPT training set designed for complex reasoning. By committing to full-scale training only after this iterative verification, we successfully transitioned the model beyond simple text completion, ensuring the internalized parameters were fully primed for the explicit logic compilation in Phase II.

\subsection{Phase II: Logic Compilation} 
To transform internalized knowledge into actionable reasoning, we utilize a specialized SFT paradigm designed to unlock the model's latent analytical capabilities. 
In order to activate the knowledge internalized during the CPT phase, we first develop a data augmentation pipeline using extremely limited QAs constructed from CPT corpus. 
Each pair is augmented with self-reflection rationales in the thinking process to assist the models in understanding their internal memorization.
We then "compile" complex business SOPs, standardized response templates and basic tool usages into the model's intuitive reasoning paths. By mixing direct answering, RAG-conditioned samples, and structured recall, we ensure the model can flexibly apply its parameters to diverse query types.


\subsection{Phase III: Alignment and Refinement} 
In this phase, we use RL to address long-tail failures, advanced function calling reasoning and reducing the robotic phrasing commonly existing in base models to make our assistant models more human-like. 
Leveraging a high-fidelity user simulator, we fine-tune the model to handle multi-turn nuances and maintain anthropomorphic resonance while strictly adhering to financial compliance.

\paragraph{Multi-objective Reward Modeling} 
\label{sec:reward_modeling}
To align model preferences with professional quality standards, we developed a multi-object reward model (RM) that evaluates agent performance across a hierarchical taxonomy of technical and behavioral dimensions. 
Table~\ref{tab:annotation_rules} shows detailed rubrics, which can decompose service excellence into granular metrics, ranging from objective SOP adherence and information security to subjective emotional resonance and linguistic consistency. 
By transitioning from a holistic scoring approach to a decomposed scoring logic, we ensure the model receives dense, multi-faceted feedback that explicitly penalizes specific failure modes—such as logical contradictions or procedural deviations—while rewarding anthropomorphic flexibility. 
Furthermore, the RM is integrated into a data flywheel architecture; this system automates bad-case attribution and facilitates real-time data accumulation to iteratively refine evaluation standards across over 100 business categories.

\paragraph{Multi-dimension User Simulation} 
To facilitate high-fidelity multi-round reinforcement learning, we implemented an inductive user simulator that reconstructs latent personas directly from authentic service logs rather than relying on abstract rule-based definitions. 
Each persona is defined through a four-dimensional attribute matrix: Background Description, Knowledge Blind Spots, Operational History, and Problem Specification. By extracting these attributes from historical human-to-human dialogues, the simulator preserves the natural distribution of user intent and the specific cognitive gaps inherent in financial service interactions.

The simulation is optimized through a two-stage training process. During the Supervised Fine-Tuning (SFT) phase, the model is trained on a mixture of filtered human dialogues and high-quality synthetic data to master diverse conversational styles, such as fragmented statements and multi-turn inquiries. This is followed by a Direct Preference Optimization (DPO) phase, which specifically penalizes repetitive or robotic response patterns identified through a rule-based discriminator, ensuring the simulator maintains a natural, human-like linguistic trajectory.

\paragraph{Multi-round Reinforcement Learning}
The RL framework optimizes multi-round dialogue trajectories, enabling the model to refine its conversational strategy over extended interactions. By integrating RAG-augmented reasoning, the framework ensures that each turn is grounded in external knowledge, utilizing closed-loop feedback with our inductive user simulator to improve long-term coherence and factual accuracy.
During each training rollout, the agent's multi-round service trajectories are evaluated by the multi-dimensional reward model established in Sec.~\ref{sec:reward_modeling}, which provides automated and high-frequency feedback across the 20-dimension evaluation hierarchy.

This integrated setup forces the model to explore complex reasoning paths and utilize its internalized knowledge to resolve inquiries before defaulting to human escalation, effectively bridging the gap between raw information recall and professional industrial reasoning.

\subsection{Expert Evaluation Framework}
\label{sec:human_eval}

To ensure that \methodname not only achieves high performance on general metrics but also meets the rigorous standards of financial customer service—specifically regarding compliance, professional accuracy, and user experience—we constructed a fine-grained \textbf{Business Expert Annotation Framework}.
This evaluation is conducted by senior Subject Matter Experts (SMEs) to rigorously audit and score model-generated responses across three core dimensions: \textit{Compliance \& Security}, \textit{Solution Effectiveness}, and \textit{Service Soft Skills}.

Our evaluation protocol adopts a ``granular error attribution'' mechanism. We decompose high-level quality indicators into specific \textbf{Error Codes} (e.g., G01 for ID verification failure, G18 for robotic tone). Each code corresponds to an evaluation item in Table~\ref{tab:annotation_rules} and is defined with explicit \textbf{Judgment Steps} and strict \textbf{Binary Scoring Rules} (0/1), where a score of 0 indicates a violation or failure, and 1 indicates compliance or success. This design minimizes subjective variance among human annotators, ensuring both consistency and objectivity in the evaluation process.

The framework specifically covers:
\begin{itemize}
    \item \textbf{Compliance Key Errors:} Critical ``veto'' items that audit information confidentiality (e.g., PII masking), identity verification processes (KYC), and multilingual consistency to ensure the baseline security of financial services.
    \item \textbf{Solution Effectiveness:} Assesses whether the model accurately grasps user intent, provides factually correct information, maintains contextual coherence, and strictly follows Standard Operating Procedures (SOPs).
    \item \textbf{Service Soft Skills:} Evaluates the anthropomorphic quality of the model, including empathy, linguistic fluency, and the ability to educate users on complex policies without being robotic.
\end{itemize}

The detailed annotation criteria and scoring logic for each Error Code are presented in Table~\ref{tab:annotation_rules}.

\section{Experiments}


\subsection{Phase I: Domain Knowledge Injection}

\paragraph{Data Construction}

\begin{figure}[htbp!]
    \centering
    \includegraphics[width=1\linewidth]{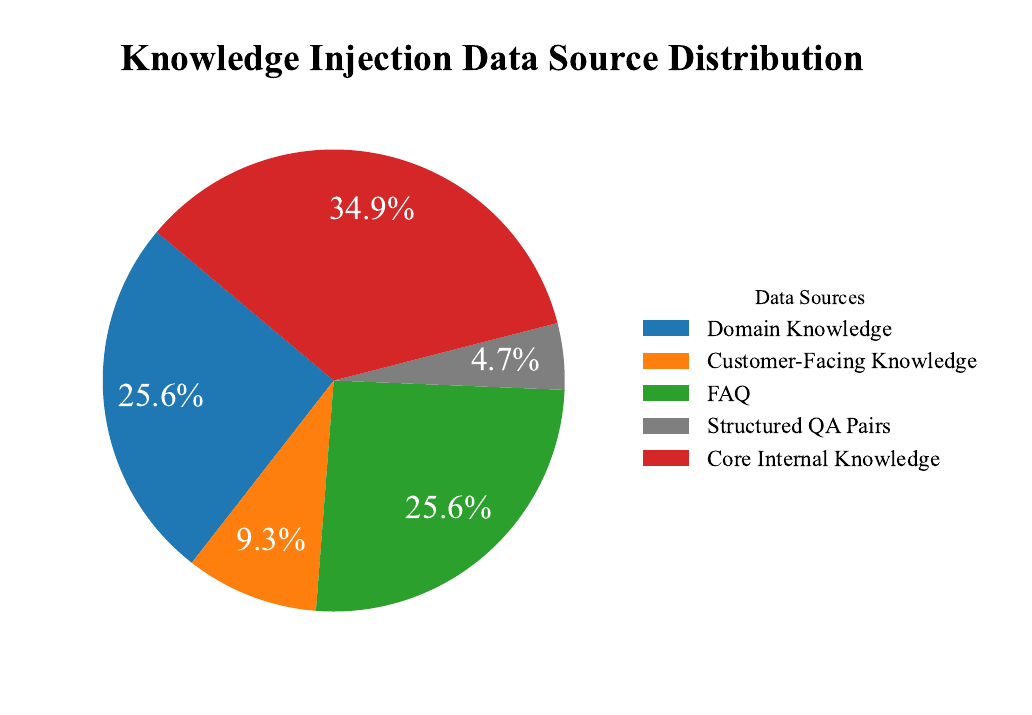}
    \caption{Best practices about the \textit{optimal sampling ratio} of data sources during knowledge injection (CPT), showing that core internal knowledge (34.9\%) constitutes the largest share, followed by domain knowledge and FAQs (both 25.6\%), with structured QA pairs representing the smallest segment (4.7\%). Note that this reflects the training-time sampling distribution rather than the raw dataset composition (detailed in Appendix~\ref{sec:cpt}). }
    \label{fig:cpt_data}
\end{figure}

The training corpus is synthesized from a diverse mix of mission-critical SOPs, financial axioms, expert interaction logs, and instructional scripts, experimentally balanced with general data to maintain linguistic versatility. 
We distill through a semantic scrubbing pipeline that leverages N-gram overlap filtering to prune administrative boilerplate and Perplexity (PPL) scrutiny via our reference model.

To enhance robustness, we implement temporal calibration with knowledge-effective timestamps and entity normalization via masking to mitigate overfitting on sensitive identifiers. 
Finally, declarative facts are refactored into deductive formats and pseudo-instructional templates, establishing a specialized latent representation and instruction-following potential during the CPT phase.

\paragraph{Evaluations}
To evaluate the effectiveness of the \methodname paradigm, we conduct experiments on our rigorous onboarding exams, Service Specialist Qualification (SSQ), required for our frontline business staff and financial experts.

\paragraph{Results}
Table~\ref{tab:phase1_results} presents the efficacy of knowledge internalization.
Our results demonstrate that \methodname (KI) achieves a score of 68.28 in a strictly parameter-only setting, being comparable with human professional proficiency (70.00) without access to notebook.
This validates the dual necessity of our paradigm: only SFT yields limited gains (60.35), confirming that CPT is the essential stepping stone for anchoring domain axioms, while SFT serves as the strategic activation trigger.

Table~\ref{tab:phase1_results_kr} further highlights the leverage effect of our Knowledge Reflection (KR) mechanism. 
By utilizing only 72 original seed kernels, \methodname achieves a significant performance leap across all model scales. 
Notably, for the Qwen3-80B-A3B variant, the addition of KR propels in-domain scores to 90.15 ($+32.79$), officially surpassing the human open-book benchmark (89.36).
Crucially, KR significantly broadens the model's generalization envelope. 

We also observe a clear scaling law in representation density: while the 8B model shows the largest raw gain in-domain, the 80B architecture maintains the highest fidelity and superior transferability in out-of-domain scenarios (75.71). This confirms that our weakly supervised distillation effectively projects sparse knowledge kernels into a robust, generalized reasoning space, meeting professional industrial standards even in long-tail scenarios.

\begin{table}[htbp]
\centering
\caption{Results of Qwen3-80B-A3B's Phase I knowledge internalization. KI = knowledge internalization. }
\label{tab:phase1_results}
\resizebox{\linewidth}{!}{
\begin{tabular}{l c c c}
\toprule
\textbf{Candidates} & \textbf{Training Methods} & \textbf{Knowledge Access} & \textbf{SSQ Score} \\ \midrule
Human Experts & - & Closed-book & 70.00 \\
Human Experts & - & Open-book & 89.36 \\\midrule
Base Model &  - & Closed-book & 56.91 \\
Base Model+KI & Only SFT & Closed-book & 60.35 \\
Base Model+KI & CPT+SFT & Closed-book & 68.28 \\ \bottomrule
\end{tabular}
}
\end{table}

\begin{table}[htbp]
\centering
\caption{Ablation study about knowledge reflection with only \textbf{72 original seed data}. KI = knowledge internalization, KR = knowledge reflection. }
\label{tab:phase1_results_kr}
\resizebox{\linewidth}{!}{
\begin{tabular}{l c c}
\toprule
\textbf{Models} & \textbf{In-domain Exams} & \textbf{Out-of-domain Exams} \\ \midrule
Qwen3-8B+KI & 49.24 & 55.36 \\
Qwen3-8B+KI+KR & 87.69 \color{red}{(+38.45)} & 71.07 \color{red}{(+15.71)}\\\midrule
Qwen3-30B-A3B+KI & 52.27 & 52.50 \\
Qwen3-30B-A3B+KI+KR & 87.50\color{red}{(+35.23)} & 76.43\color{red}{(+23.93)} \\\midrule
Qwen3-80B-A3B+KI & 57.36 & 56.07 \\
Qwen3-80B-A3B+KI+KR & 90.15\color{red}{(+32.79)} & 75.71\color{red}{(+19.64)} \\
\bottomrule
\end{tabular}
}
\end{table}

\subsection{Phase II: Business Logic Compilation}

\paragraph{Data Construction} 
The final model was trained on a meticulously curated 200k instruction-tuning set, designed to bridge the gap between static domain axioms and dynamic service interactions. This corpus is anchored by high-fidelity exemplar dialogues derived from premium historical service logs and procedural QA pairs refactored from internal SOPs. To maintain the model's core reasoning potential and linguistic versatility, we further integrated a 30\% mixture of Phase I SFT data and general-purpose instruction-following samples. This balanced composition ensures that \methodname internalizes specialized business logic while preserving its generalized instruction-following and self-reflection capabilities.

\paragraph{Results} 
Table~\ref{tab:final_robustness} illustrates the performance of the integrated \methodname pipeline. In a strictly parameter-only setting, our 8B candidate achieves a score of 91.96, surpassing the professional human experts with or without knowledge access. 
This significant improvement from the 49.13 baseline demonstrates that the strategic combination of Phase I and Phase II successfully internalizes mission-critical domain knowledge into the model's weights.

Notably, \methodname without any context outperforms the massive Qwen3-235B teacher model operating in RAG capacity (91.71). These results validate that our hierarchical knowledge management and multi-stage internalization paradigm effectively bridge the gap between static domain and actionable reasoning, meeting professional standards in high-stakes merchant service scenarios.

\begin{table}[htbp]
\centering
\caption{Final evaluation across the SSQ Exam and held-out Closed Sets. \methodname matches or exceeds both human experts and massive teacher models in a parameter-only setting. PI=Phase I, PII=Phase II. }
\label{tab:final_robustness}
\small
\resizebox{1\linewidth}{!}{ 
\begin{tabular}{l c c}
\toprule
\textbf{Candidate} & \textbf{Knowledge Access} & \textbf{SSQ Score}\\ \midrule
Human Experts  & Closed-book & 70.00 \\
Human Experts  & Open-book & 89.36 \\\midrule
Qwen3-8B-Instruct & Open-book & 82.43 \\
Qwen3-30B-A3B & Open-book & 88.86 \\
Qwen3-235B-A22B & Open-book & 91.71  \\ \midrule
Qwen3-8B-Instruct & Closed-book & 49.13 \\
Qwen3-30B-A3B & Closed-book & 52.35 \\
\midrule
Qwen3-8B-Instruct+PI+PII & Closed-book & \textbf{91.96}  \\ \bottomrule
\end{tabular}
}
\end{table}



\subsection{Phase III: Alignment and Refinement}

\paragraph{User Simulator Evaluation}
We assess our user simulator through a human indistinguishability evaluation. We mixed 1,000 samples of dialogue between the user simulator and the assistant model with real human-assistant interactions and presented them to customer service experts. Annotators were asked to classify the user's responses as "Human", "Robot", or "Uncertain". The \textit{Human-likeness} is defined as the proportion of samples labeled as "Human" or "Uncertain". 
As shown in Table~\ref{tab:user_sim_eval}, our user simulator achieves a Human-likeness of 95.8\%, closely mirroring the 96.7\% score observed for real humans.

\begin{table}[htbp]
\centering
\caption{User Simulator Human-likeness Evaluation Results. Our user simulator achieves human-level fidelity.}
\label{tab:user_sim_eval}
\small
\begin{tabular}{lc}
\toprule
\textbf{Role} & \textbf{Human-likeness} \\ \midrule
Real Human & 96.7\% \\
User Simulator & 95.8\% \\
\bottomrule
\end{tabular}
\end{table}

\paragraph{Data Construction} 
The multi-turn reinforcement learning dataset consists of 3,000 unique user simulator profiles that span a wide range of business scenarios. Each profile is then used by a user simulator model to emulate a user during multi-turn trajectory with the policy model during rollout.

\paragraph{Evaluations} 
We evaluate on a held-out test suite of real-world, multi-round service scenarios, each paired with a structured user profile specifying goals and context to drive simulated user–assistant interactions. We report the KGA (Knowledge-Grounded Aggregate) score—a trajectory-level scalar computed as a weighted average of the 20 scoring dimensions in Sec.~\ref{sec:reward_modeling}, with higher weights assigned to knowledge-grounding dimensions. The same KGA metric is used for training-time reward computation and test-time evaluation, ensuring  alignment across the pipeline.

\paragraph{Results} 
\methodname RL delivers substantial gains in KGA. As shown in Table~\ref{tab:rl_eval}, the 8B model improves from 81.41 to 95.05 (+13.64), and the 32B model from 85.11 to 96.13 (+11.02). Scaling from 8B to 32B helps the base model (+3.70), while the post-RL gap is modest (+1.08), showing reduced reliance on model size. Both RL variants exceed 95 KGA, indicating that RL is the primary driver of the grounding and coherence benefits.

\begin{table}[htbp]
\centering
\small
\caption{RL evaluation. \methodname shows robust improvements over base model performance across 8B/32B model scales.}
\label{tab:rl_eval}
\begin{tabular}{l c c}
\toprule
\textbf{Model} & \textbf{KGA Score}\\ \midrule
Qwen3-8B-Instruct+PI+PII  &  81.41 \\
Qwen3-8B-Instruct+PI+PII+PIII &  95.05 \\ \midrule
Qwen3-32B-Instruct+PI+PII  & 85.11 \\
Qwen3-32B-Instruct+PI+PII+PIII & 96.13 \\
\bottomrule
\end{tabular}
\end{table}

\section{Conclusions}
We presented \methodname, a paradigm that shifts industrial agent architecture from brittle, fragmented workflows to internalized cognitive intuition. 
By implementing a Hierarchical Knowledge Management strategy through a progressive multistage pipeline, we successfully consolidated complex financial SOPs and domain axioms into a unified attention space. 
Extensive real-world deployment confirms that this single-model approach effectively breaks the industrial trade-off between latency, accuracy, and complexity: achieving a 57.2\% reduction in latency while simultaneously elevating Intent Resolution Rates to 83.3\% and doubling reasoning performance on complex cases.
Ultimately, \methodname demonstrates that the future of high-stakes enterprise AI lies not in orchestrating external modules, but in evolving domain-specialized foundational models, offering a scalable, robust, and human-centric blueprint for the next generation of domain-specialized agents.

\section*{Limitations and Future Work}

The transition from a modular pipeline to \methodname yielded several non-trivial insights. These limitations highlight the bitter lessons between academic metrics and industrial viability in high-stakes financial environments.\paragraph{1. Style Fidelity $\neq$ Behavioral Fidelity (The Turing Trap)}In our User Simulator experiments, we initially celebrated a 95.8\% Human-likeness score in human indistinguishability evaluations, believing our simulator perfectly mirrored real users. However, online A/B testing revealed a shocking disconnect: the Kappa coefficient between the simulator's "satisfaction" and real users' behavior was merely 0.0446 (near random).We discovered that while the simulator sounded like a human (style), it did not act like a financial customer (behavior). Specifically, it failed to exhibit "irrational persistence" or the urge to "escalate to human agents" (Real Transfer Rate: $\approx10\%$ vs. Simulator: $1.7\%$).The lesson is that linguistic fluency is a superficial metric for user simulation. To train a robust RL policy, the simulator must be explicitly trained on \textit{negative behaviors}—impatience, complaints, and the demand for human intervention—rather than just politeness.

\paragraph{2. Reward Modeling is Data Cleaning, Not Just Training}A major bottleneck in our RL phase was the high inconsistency in human annotation. We found that simply training a Reward Model (RM) on noisy labels led to policy degradation.We shifted our paradigm from "training an RM" to "building a data-cleaning flywheel." We established a closed-loop system where an iterative RM was used to filter and correct its own training data (identifying "bad cases" in human labels).We learned that in industrial RLHF, the quality of the Reward Model is defined less by its architecture and more by the hygiene of its data pipeline. A weaker model trained on rigorous, model-assisted clean data consistently outperformed larger models trained on raw human annotations.\paragraph{3. Global Optimization Over Modular Debuggability}Traditional engineering wisdom favors modular "Agentic Workflows" for their interpretability and fault isolation. However, we found that the rigid state-machine transitions in PEER architectures caused severe "Semantic Discontinuities"—information loss during module handoffs led to logic jumps that no single module could resolve.By switching to \methodname, we sacrificed the "comfort" of white-box debugging (knowing exactly which module failed) for the "performance" of a unified attention space.The lesson is that local optimization (improving a specific Router or Retriever) often leads to global sub-optimization. Only a unified model, where understanding, reasoning, and expression evolve synchronously, can achieve the semantic coherence required for complex financial inquiries.\paragraph{4. The High Cost of Knowledge Pollution}In early iterations, we aggressively injected all business rules into the model via Continual Pre-training (CPT). We found that volatile parameters (e.g., campaign rules, exchange rates) became "toxic knowledge" that was impossible to update without catastrophic forgetting.We established a strict Hierarchical Knowledge Governance: CPT is reserved \textit{only} for immutable Domain Axioms (the "Physics" of finance). Semi-static Procedural Logic is delegated to SFT (Instruction Following), while volatile data must remain external.We learned that model parameters are a premium storage medium, not a garbage bin. Misusing CPT for transient data creates a "high-maintenance debt" that outweighs any reasoning gains.

\paragraph{5. The Helpfulness vs. Compliance Conflict}
In general-purpose LLMs, RLHF typically rewards "helpfulness." In financial services, however, an agent that is "too helpful" (e.g., bypassing KYC to please a frustrated user) poses a severe compliance risk. We observed that standard RL training naturally drifted towards sycophancy—the model would agree with user errors to maximize immediate satisfaction rewards. To counter this, we had to shape principled refusal into the reward function. 
We learned an industrial agent must know when to be unhelpful to reduce hallucinations. 
Balancing "Task Success" with "Regulatory Adherence" requires a multi-objective reward system where compliance violations act as a hard veto.

Building on these bitter lessons, our future roadmap focuses on three strategic frontiers to perfect the \methodname paradigm:

\begin{itemize}
    \item \textbf{From Static Replay to Adversarial Self-Play:} 
    To escape the "Turing Trap," we will transition from log-replay simulators to \textit{Adversarial World Models}. By employing \textbf{Asymmetric Self-Play}, we intend to explicitly reward the User Simulator for successfully inducing compliance violations or exposing logic loops in the Agent. This evolutionary pressure will force the Agent to master "principled refusal" and resilience against irrational persistence, effectively turning rare corner cases into the training distribution.
    
    \item \textbf{SOP-Aligned Process Supervision (PRMs):} 
    Addressing the "Black-box" challenge requires more than outcome-based rewards. We aim to integrate Fine-grained Process Reward Models to scrutinize the model's reasoning trajectory at each intermediate step. By administering dense reward signals for every valid logical transition, we can implicitly reinforce the structural integrity of sequential \textit{Chain-of-SOP} reasoning and precise tool invocation. This ensures structured and interpretable reasoning without sacrificing the fluency of a unified model.
    
    \item \textbf{Surgical Model Editing for Volatile Knowledge:} 
    To resolve "Knowledge Pollution," we will explore \textbf{Rank-One Model Editing (ROME)} and parametric memory separation. Our goal is to dissociate volatile data (e.g., daily exchange rate spreads, flash campaign dates) from stable reasoning weights. This will enable minute-level, surgical updates to specific factual associations directly within the parameters, preventing catastrophic forgetting of the broader dialogue policies.
\end{itemize}

\section*{Acknowledgements}

Our work has benefited immensely from the extensive support of several teams at Ant International. We would like to express our sincere gratitude to the business, product design, engineering, and quality assurance teams for their significant contributions. They provided invaluable business insights, strategic guidance, and constructive suggestions that shaped the direction of this research. Furthermore, their technical expertise in system integration and rigorous quality validation was essential to the successful implementation of our models.

\section*{Contributions}

Chang Liu, Chaoyang Ning, Dayi Jiang, Enrui Gu, Fang Ran,
Hongyan Xue, Huaqing Li, Hui Cai, Jia Liu, Jiang-Ming Yang, Jianshe Li, Jiawei Luo, Jin Zhou, Leshen Zhu, Lihui Chen,
Liying Ma, Lyuxin Xue, Mengjian Ji, Ruijia Xu, Wei Ren, Wei Wu, Xiaoling Qu,
Xiaoyun Feng, Xin Zhang, Xixie Zhou, Xuanwei Hu, Yan Chen, Yichao Wang, Yongqi Tong,
Yu Liu, Yuhong Zhou, Zemin Sun, Zhenwen Xu, Zhiling Liu, Zifan Wang

\bibliography{custom}

\clearpage
\appendix

\section{Related Work}

\paragraph{Business Knowledge Management and Injection}

The management and injection of specialized business knowledge have evolved from fragmented modular workflows to hierarchical internalization strategies that optimize the trade-off between parameter rigidity and operational fluidity. As demonstrated in the OneModel Paradigm, business intelligence is structured into a three-tier hierarchy: Fundamental Domain Axioms (static principles), Procedural Business Logic (semi-static SOPs), and Volatile Transactional Context (dynamic real-time data). To internalize this expertise, the framework utilizes a Multi-stage Knowledge Injection pipeline. Phase I: Knowledge Injection leverages Continual Pre-training (CPT) on massive financial and e-commerce corpora to establish a robust "closed-book" semantic foundation, aligning with recent trends in scaling domain-specific proficiency \citep{kimiteam2025kimik15scalingreinforcement, deepseekai2025deepseekr1incentivizingreasoningcapability, tong2024optimizinglanguagemodelsreasoning,pan-etal-2025-understanding}. This is followed by Phase II: Logic Compilation, which uses Supervised Fine-Tuning (SFT) to transform internalized facts into actionable reasoning paths by training on complex SOPs and response templates. This comprehensive injection process serves as a necessary prerequisite for subsequent reinforcement learning, ensuring the model can maintain strategic coherence and "thinking" depth when faced with the underspecified or contextually fragmented inputs often encountered in professional service environments \citep{wang2025bpobalancedpreferenceoptimization, he2025rethinkingreasoningqualitylarge, chu2025sftmemorizesrlgeneralizes, yeo2025demystifyinglongchainofthoughtreasoning,chen2025dingtalkdeepresearchunifiedmulti,tong2024optimizinglanguagemodelsreasoning,wang2023planandsolvepromptingimprovingzeroshot,lin2023toxicchat}.

\paragraph{User Simulation}
Traditional user simulators, once constrained by rule-based~\cite{schatzmann2007agenda} or supervised designs~\cite{lin2021domain}, have been largely superseded by LLM-based paradigms that offer superior generalization. In general and role-playing contexts, these models typically utilize zero-shot prompting~\cite{xu2023baize,ding2023enhancing}, targeted fine-tuning~\cite{kong2024platolm,sun2024parrot}, or persona-driven constraints~\cite{shao2023character,wang2025know} to effectively mimic diverse human behaviors. For Task-Oriented Dialogue (TOD), where maintaining goal consistency is paramount, recent approaches have integrated validation mechanisms~\cite{luo2024duetsim} or domain-specific fine-tuning~\cite{sekulic2024reliable} to minimize hallucinations. Moving toward agentic workflows, frameworks like $\tau^2$-Bench~\cite{barres2025tau} explicitly couple user actions with environmental tools to ensure strict state adherence. Compared to traditional TOD systems, task-oriented agent environments introduce greater realism and complexity, thereby imposing stricter requirements on a simulator's alignment with authentic human trajectories.

\paragraph{Reinforcement Learning}
Reinforcement Learning (RL) has emerged as a transformative paradigm for scaling the general proficiency and interactive capabilities of LLMs \citep{kimiteam2025kimik15scalingreinforcement,deepseekai2025deepseekr1incentivizingreasoningcapability,tong2024optimizinglanguagemodelsreasoning,wang2025bpobalancedpreferenceoptimization,he2025rethinkingreasoningqualitylarge,xu2025rewardconsistencyimprovingmultiobjective}. By navigating a vast space of potential interaction trajectories, RL enables the policy to move beyond its initial training distribution and discover sophisticated behavioral patterns and novel problem-solving strategies \citep{chu2025sftmemorizesrlgeneralizes, yeo2025demystifyinglongchainofthoughtreasoning,pan-etal-2025-understanding}. The efficacy of this optimization is underpinned by diverse reward mechanisms—ranging from final outcome verifiers (outcome supervision) to process-level preference models (process supervision)—which provide a scalable target for aligning model outputs with complex professional standards \citep{lambert2024tulu, lightman2023letsverifystepstep, uesato2022solvingmathwordproblems}. However, while these methods have shown success in well-defined tasks, they are often applied in idealized, single-turn settings. This leaves a significant gap in multi-round reinforcement learning, particularly regarding the challenge of maintaining robust reasoning and strategic coherence when faced with the real-world complexity of underspecified queries, contradictory premises, or contextually fragmented dialogues.

\section{Model Serving and Deployment}
\label{sec:deployment}

Our unified \methodname models are deployed on our large-scale internal \platform infrastructure, leveraging state-of-the-art hardware and serving frameworks to ensure high-throughput and low-latency inference.
Specifically, we utilize NVIDIA H200 Tensor Core GPUs, which offer 141GB of HBM3e memory and 4.8TB/s bandwidth, significantly alleviating the memory-bound bottlenecks typical in large-scale LLM serving.

For the inference backend, we employ the vLLM library~\citep{kwon2023efficientmemorymanagementlarge}, a high-performance serving engine designed to optimize memory utilization and generation speed. 
To ensure seamless integration with our upstream applications, the models are exposed via vLLM's OpenAI-compatible API server.
System reliability and resource utilization (e.g., GPU memory usage, Time Per Output Token) are continuously monitored through our internal observability dashboards, with auto-scaling policies configured to handle diurnal traffic patterns dynamically.

\section{Training Hyper-parameters}

\begin{table}[htbp!]
    \centering
    \caption{Phase I's hyperparameters implemented based on the swift framework.}
    \label{tab:sft_hyperparams}
    
    \resizebox{0.9\linewidth}{!}{ 
    \begin{tabular}{w{c}{3.5cm}w{c}{5cm}w{c}{5cm}}
    \toprule
         \textbf{Category} &  \textbf{Hyperparameter} & \textbf{Value} \\ 
         \midrule
         \multirow{4}{*}{Trainer} &  Nodes & 8 ($nnodes$)\\
                                  &  GPUs per node & 8\\
                                  &  Num train epochs & 10\\
                                  &  Deepspeed stage & zero3\\
         \midrule
         \multirow{5}{*}{Model} 
                                &  Max length & 8192\\
                                &  Attention impl & flash\_attn\\
                                &  Torch dtype & bfloat16\\
                                &  Padding free & True\\
                                &  Packing & True\\
         \midrule
         \multirow{5}{*}{Optimization} &  Learning rate & $1\times10^{-4}$\\
                                       &  Per-device batch size & 1\\
                                       &  Gradient accumulation steps & 16\\
                                       &  Warmup ratio & 0.1\\
                                       &  Optimizer & Adam \\
         \midrule
         \multirow{5}{*}{Logging / Save} &  Eval steps & 100\\
                                         &  Save steps & 200\\
                                         &  Save total limit & 10\\
                                         &  Save only model & True\\
                                         &  Logging steps & 10\\
         \bottomrule
    \end{tabular}
    }
\end{table}

\begin{table}[htbp!]
    \centering
    \caption{Hyperparameters for Phase II (Logic Compilation via SFT). The model is fine-tuned on complex business SOPs using the SWIFT framework.}
    \label{tab:sft_hyperparams_phase2}
    
    \resizebox{0.9\linewidth}{!}{ 
    \begin{tabular}{l l c}
    \toprule
    \textbf{Category} & \textbf{Hyperparameter} & \textbf{Value} \\ 
    \midrule
    \multirow{2}{*}{Infrastructure} & Framework & SWIFT \\
                                    & Number of GPUs & 8 \\
    \midrule
    \multirow{6}{*}{Optimization}   & Learning Rate & $3\times10^{-5}$ \\
                                    & Per-Device Batch Size & 4 \\
                                    & Epochs & 2 \\
                                    & Optimizer & AdamW \\
                                    & LR Scheduler & Cosine \\
                                    & Warmup Ratio & 0.05 \\
    \midrule
    \multirow{1}{*}{Tokenization}   & Max Sequence Length & 8,192 \\
    \bottomrule
    \end{tabular}
    }
\end{table}

\begin{table}[htbp!]
        \centering
        \caption{Phase III's hyperparameters implemented based on the veRL framework.}
        \label{tab:ppo_hyperparams}
        
        \resizebox{0.9\linewidth}{!}{ 
        \begin{tabular}{w{c}{3.5cm}w{c}{5cm}w{c}{5cm}}
        \toprule
             \textbf{Category} &  \textbf{Hyperparameter} & \textbf{Value} \\ 
             \midrule
             \multirow{5}{*}{Trainer} &  Nodes & 1\\
                                      &  GPUs per node & 8\\
                                      &  Total steps & 726\\
                                      &  Gradient checkpointing & True \\
                                      &  Use remove padding & True \\
             \midrule
             \multirow{2}{*}{Algorithm} &  Advantage estimator & GRPO \\
                                        &  Use KL in reward & False\\
             \midrule
             \multirow{7}{*}{Actor} &  Learning rate & $1\times10^{-6}$ \\
                                    &  Train batch size & 4 \\ 
                                    &  Mini-batch size & 64 \\ 
                                    &  Ulysses sequence parallel size & 2 \\ 
                                    &  Use KL loss & True (coeff=0.001) \\ 
                                    &  Optimizer warmup & Cosine \\ 
                                    &  LR warmup steps ratio & 0.05 \\ 
             \midrule
             \multirow{7}{*}{Rollout} &  Backend & vLLM \\
                                      &  Mode & Async \\ 
                                      &  Rollout $n$ & 8 \\ 
                                      &  Max turns & 5 \\
                                      &  User Sim (Temp/Top-p/Top-k) & 0.6 / 0.95 / 20 \\ 
                                      &  Assistant (Temp/Top-p/Top-k) & 1.0 / 1.0 / -1 \\ 
                                      &  Tensor model parallel size & 4 \\ 
             \midrule
             \multirow{1}{*}{Reward Model} &  Backend & vLLM \\ 
             \bottomrule
        \end{tabular}
        }
\end{table}

\begin{table}[htbp!]
    \centering
    \caption{Hyperparameters for User Simulator Training in Phase III (Alignment and Refinement). Both stages are implemented using the SWIFT framework on 8 GPUs.}
    \label{tab:sft_dpo_hyperparams}
    
    \resizebox{0.9\linewidth}{!}{ 
    \begin{tabular}{l l c c}
    \toprule
    \textbf{Category} & \textbf{Hyperparameter} & \textbf{SFT (Phase II)} & \textbf{DPO (Phase III)} \\ 
    \midrule
    \multirow{2}{*}{Infrastructure} & Framework & SWIFT & SWIFT \\
                                    & Number of GPUs & 8 & 8 \\
    \midrule
    \multirow{6}{*}{Optimization}   & Learning Rate & $3\times10^{-5}$ & $3\times10^{-5}$ \\
                                    & Per-Device Batch Size & 4 & 8 \\
                                    & Epochs & 2 & 2 \\
                                    & Optimizer & AdamW & AdamW \\
                                    & LR Scheduler & Cosine & Cosine \\
                                    & Warmup Ratio & 0.05 & 0.05 \\
    \midrule
    \multirow{1}{*}{Tokenization}   & Max Sequence Length & 8,192 & 20,000 \\
    \midrule
    \multirow{2}{*}{Algo. Specific} & KL Penalty ($\beta$) & - & 0.1 \\
                                    & RPO Alpha ($\alpha$) & - & 0.1 \\
    \bottomrule
    \end{tabular}
    }
\end{table}

\section{CPT Data Construction and Ablation for Phase I: Knowledge Injection}
\label{sec:cpt}
To ensure deep domain alignment without compromising general capabilities, we constructed a high-density financial corpus comprising three strategic tiers, distributed to balance internal logic with external applicability.First, Core Internal Knowledge (approx. 60\% of the corpus): This dominant tier consists of high-fidelity internal training manuals and augmented documents, serving as the "ground truth" for business axioms and implicit logic.Second, Customer-Facing Knowledge ($\sim$15\%): This segment includes a comprehensive collection of product documentation and official website guides, representing the standard operating procedures accessible to end-users.Third, Structured QA Pairs ($\sim$25\%): We curated a subset of RAG-retrieved FAQ pairs to bridge the gap between declarative knowledge and interrogative formats.All data underwent heuristic cleaning to remove HTML tags and visual noise while preserving semantic structure via document chunking. Crucially, to address the scarcity of high-value internal axioms compared to general corpora, we applied a Dynamic Up-sampling Strategy. Specifically, core training manuals were heavily up-weighted to force "rote memorization" of immutable domain rules.Ablation studies on the base model demonstrated the critical efficacy of this approach: the targeted knowledge injection yielded a significant performance leap, raising the zero-shot accuracy on core business exams by over 20 percentage points (from a baseline of $\sim$62\% to $>$82\%). This result validates our hypothesis that high-frequency repetition of low-volume, high-density domain data is a prerequisite for effective logic compilation.
Figure~\ref{fig:cpt_data} denotes our best practices of the proportion of knowledge injection data.

\section{Expert Human-in-the-Loop Annotation Rules}
\label{sec:annotation_rules_appendix}

\begin{table*}[h!]
\centering
\small
\resizebox{\textwidth}{!}{%
\begin{tabular}{p{0.12\textwidth} p{0.15\textwidth} p{0.25\textwidth} p{0.4\textwidth}}
\toprule
\textbf{Dimension} & \textbf{Sub-category}  & \textbf{Evaluation Item} & \textbf{Definition \& Scoring Logic (0 = Fail, 1 = Pass)} \\
\midrule
\multicolumn{4}{l}{\textit{\textbf{I. Compliance Key Errors (Critical Safety Layer)}}} \\
\midrule
\textbf{Privacy \& Security}
 & Identity Verification  & \textbf{ID Confirmation} &
 \textbf{Def:} Must not reveal account info without valid identity verification. \newline
 \textbf{0:} Revealed info before KYC. \textbf{1:} Followed proper protocol. \\
 \cmidrule{2-4}
 & Internal Confidentiality  & \textbf{Internal Material} &
 \textbf{Def:} Must not send internal-only materials to clients. \newline
 \textbf{0:} Leaked internal docs. \textbf{1:} Public info only. \\
 \cmidrule{2-4}
 & PII Protection  & \textbf{Data Masking} &
 \textbf{Def:} Sensitive info (e.g., phone/account numbers) must be masked. \newline
 \textbf{0:} Sent raw PII. \textbf{1:} Properly masked (e.g., 138****1234). \\
\midrule
\textbf{Attitude \& Norms}
 & Service Attitude & \textbf{Emotional Propriety} &
 \textbf{Def:} No offense/mockery; timely appeasement during conflicts. \newline
 \textbf{0:} Rude/Argumentative. \textbf{1:} Professional \& Empathetic. \\
 \cmidrule{2-4}
 & Language Consistency  & \textbf{Language Match} &
 \textbf{Def:} Reply language must match the user's inquiry language. \newline
 \textbf{0:} Mismatched language. \textbf{1:} Consistent language. \\
 \cmidrule{2-4}
 & Promise Management & \textbf{Over-commitment} &
 \textbf{Def:} AI must not promise offline actions it cannot perform (e.g., ``I will call you''). \newline
 \textbf{0:} Made empty promises. \textbf{1:} No over-commitment. \\
\midrule
\multicolumn{4}{l}{\textit{\textbf{II. Solution Effectiveness (Business Logic Layer)}}} \\
\midrule
\textbf{Accuracy}
 & Business Solution  & \textbf{Relevance} &
 \textbf{Def:} Directly address the core issue without misleading info. \newline
 \textbf{0:} Irrelevant/Misleading. \textbf{1:} Accurate \& Relevant. \\
 \cmidrule{2-4}
 & Conciseness  & \textbf{Brevity} &
 \textbf{Def:} Avoid redundant information that dilutes the solution. \newline
 \textbf{0:} Verbose/Distracting. \textbf{1:} Concise. \\
\midrule
\textbf{Intent \& Fact}
 & Intent Recognition  & \textbf{Intent Coverage} &
 \textbf{Def:} Fully cover user's explicit and implicit needs. \newline
 \textbf{0:} Missed core/sub-intents. \textbf{1:} Fully understood. \\
 \cmidrule{2-4}
 & Factual Correctness  & \textbf{Factuality} &
 \textbf{Def:} Information aligns with company policy/product facts. \newline
 \textbf{0:} Factual errors. \textbf{1:} Factually correct. \\
 \cmidrule{2-4}
 & Consistency  & \textbf{Self-Consistency} &
 \textbf{Def:} No conflicting statements within the same session. \newline
 \textbf{0:} Contradictory logic. \textbf{1:} Self-consistent. \\
\midrule
\textbf{Process \& Context}
 & Topic Adherence  & \textbf{Topic Switching} &
 \textbf{Def:} Follow user's topic switches; strictly handle business-related queries. \newline
 \textbf{0:} Failed to switch/Stuck. \textbf{1:} Adaptive switching. \\
 \cmidrule{2-4}
 & SOP Execution  & \textbf{Effective Progression} &
 \textbf{Def:} Strictly follow SOPs to collect necessary info/execute steps. \newline
 \textbf{0:} Violated SOP/Missing info. \textbf{1:} SOP compliant. \\
 \cmidrule{2-4}
 & Memory  & \textbf{Context Retention} &
 \textbf{Def:} Utilize history to avoid repeating questions. \newline
 \textbf{0:} Amnesic/Repetitive. \textbf{1:} Coherent memory. \\
\midrule
\multicolumn{4}{l}{\textit{\textbf{III. Service Soft Skills (Human-Alignment Layer)}}} \\
\midrule
\textbf{Empathy \& Style}
 & Client Education  & \textbf{Cognitive Correction} &
 \textbf{Def:} Politely correct user misconceptions and explain policies. \newline
 \textbf{0:} Blunt correction. \textbf{1:} Educational \& Polite. \\
 \cmidrule{2-4}
 & Courtesy & \textbf{Politeness} &
 \textbf{Def:} Use soft tones and appropriate honorifics; avoid robotic tone. \newline
 \textbf{0:} Cold/Robotic. \textbf{1:} Warm/Human-like. \\
 \cmidrule{2-4}
 & Fluency  & \textbf{Linguistic Quality} &
 \textbf{Def:} Clear logic and fluent expression; smooth emotional management. \newline
 \textbf{0:} Stiff/Incoherent. \textbf{1:} Fluent \& Smooth. \\
\bottomrule
\end{tabular}%
}
\caption{The Expert Human-in-the-Loop Annotation Rules. This table details the binary scoring criteria used by Subject Matter Experts (SMEs) to evaluate the model across Compliance, Solution Quality, and Soft Skills. Items unrelated to AI generation (e.g., system UI operations) are excluded.}
\label{tab:annotation_rules}
\end{table*}

\end{document}